\documentclass[manuscript]{article} 
\usepackage{aaai25}  
\usepackage{times}  
\usepackage{helvet}  
\usepackage{courier}  
\usepackage[hyphens]{url}  
\usepackage{graphicx} 
\usepackage{natbib}  
\usepackage{caption} 
\usepackage{algorithm}
\usepackage{algorithmic}
\usepackage{booktabs} 
\usepackage{amsmath}

\usepackage{newfloat}
\usepackage{listings}
\DeclareCaptionStyle{ruled}{labelfont=normalfont,labelsep=colon,strut=off} 
\floatstyle{ruled}
\newfloat{listing}{tb}{lst}{}
\floatname{listing}{Listing}
\usepackage{graphicx}
\usepackage{multirow}
\usepackage{tabularx}
\usepackage{array}
\usepackage{lscape}
\usepackage{cuted}
\usepackage{amssymb}
\usepackage{subcaption}
\usepackage{makecell}
\usepackage[table]{xcolor}
\usepackage{multirow}
\usepackage{ragged2e} 
\usepackage{booktabs,makecell, multirow, tabularx}
\usepackage{colortbl}

\title{UrBench: A Comprehensive Benchmark for Evaluating Large Multimodal Models in Multi-View Urban Scenarios}
\author{
    Baichuan Zhou\textsuperscript{\rm 1*},
    Haote Yang\textsuperscript{\rm 1*},
    Dairong Chen\textsuperscript{\rm 4,2*},
    Junyan Ye\textsuperscript{\rm 2,1}\thanks{These authors contributed equally.},\\
    Tianyi Bai\textsuperscript{\rm 1},
    Jinhua Yu\textsuperscript{\rm 2},
    Songyang Zhang\textsuperscript{\rm 1},
    Dahua Lin\textsuperscript{\rm 1,3}, \\
    Conghui He\textsuperscript{\rm 1,3†},
    Weijia Li\textsuperscript{\rm 2}\thanks{W. Li and C. He are the corresponding authors.}
}
\affiliations{
    \textsuperscript{\rm 1}Shanghai AI Laboratory \\
    \textsuperscript{\rm 2}Sun Yat-Sen University \\ 
    \textsuperscript{\rm 3}Sensetime Research \\
    \textsuperscript{\rm 4}Wuhan University \\
    \{chendr7, yejy53, yujh56\}@mail2.sysu.edu.cn, 
    liweij29@mail.sysu.edu.cn, \\
    \{zhoubaichuan, yanghaote, baitianyi, zhangsongyang, lindahua, heconghui\}@pjlab.org.cn
}

\usepackage{bibentry}

\begin{document}

\maketitle

\begin{abstract}
Recent evaluations of Large Multimodal Models (LMMs) have explored their capabilities in various domains, with only few benchmarks specifically focusing on urban environments. Moreover, existing urban benchmarks have been limited to evaluating LMMs with basic region-level urban tasks under singular views, leading to incomplete evaluations of LMMs' abilities in urban environments.
To address these issues, we present UrBench, a comprehensive benchmark designed for evaluating LMMs in complex multi-view urban scenarios. 
UrBench contains 11.6K meticulously curated questions at both region-level and role-level that cover 4 task dimensions: Geo-Localization, Scene Reasoning, Scene Understanding, and Object Understanding, totaling 14 task types. 
In constructing UrBench, we utilize data from existing datasets and additionally collect data from 11 cities, creating new annotations using a cross-view detection-matching method. With these images and annotations, we then integrate LMM-based, rule-based, and human-based methods to construct large-scale high-quality questions.
Our evaluations on 21 LMMs show that current LMMs struggle in the urban environments in several aspects. Even the best performing GPT-4o lags behind humans in most tasks, ranging from simple tasks such as counting to complex tasks such as orientation, localization and object attribute recognition, with an average performance gap of 17.4\%. 
Our benchmark also reveals that LMMs exhibit inconsistent behaviors with different urban views, especially with respect to understanding cross-view relations.
\end{abstract}

\begin{links}
    \link{Project}{https://opendatalab.github.io/UrBench/}
    \link{Appendix}{https://github.com/opendatalab/UrBench/blob/master/static/appendix.pdf}
\end{links}

\section{Introduction}\label{sec:intro}

\begin{figure}[t]
    \centering
    \includegraphics{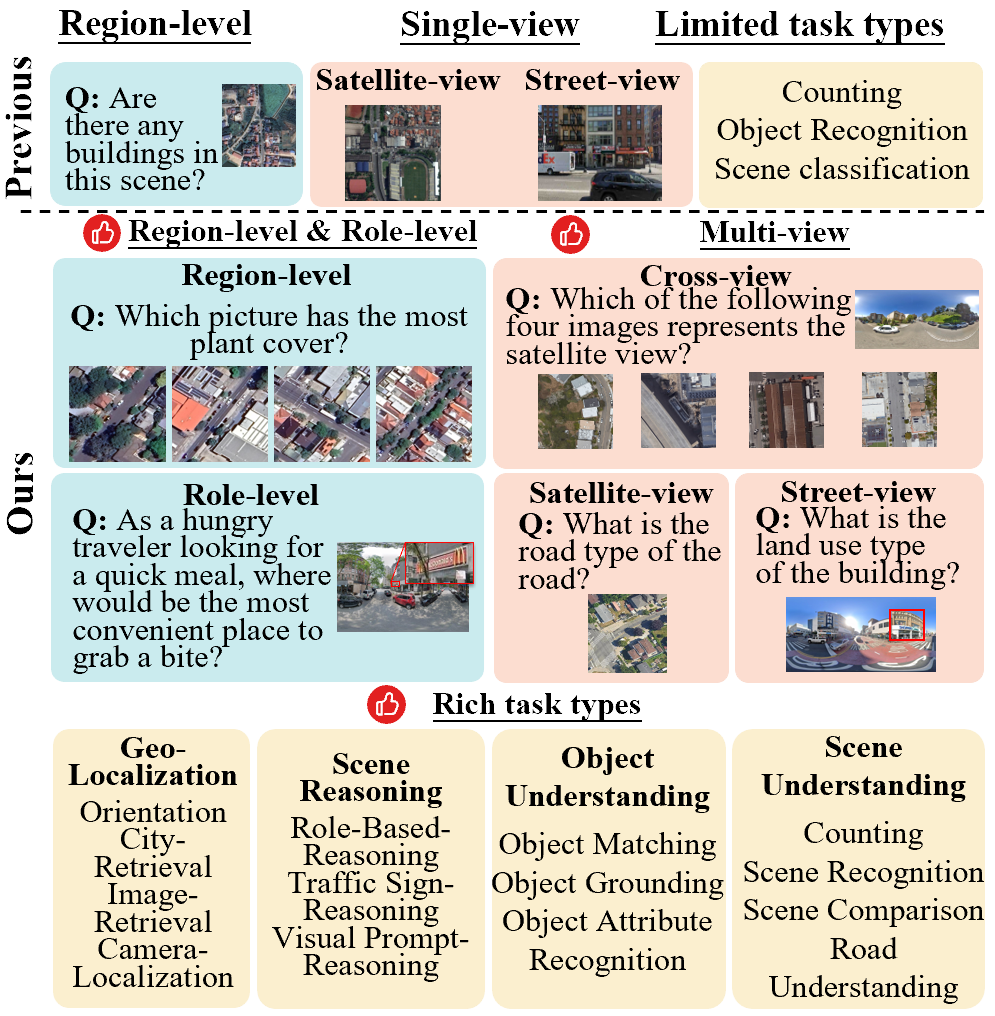}
    \caption{Comparison between UrBench and previous works. (1) UrBench contains both region-level and role-level questions, while previous benchmarks generally focus on region-level questions. (2) In addition to single-view questions in satellite or street view, UrBench also incorporates cross-view questions. (3) It evaluates LMMs on a comprehensive range of 14 diverse tasks in 4 evaluation dimensions.}
    \label{fig:question-comparison}
\end{figure}

Recently, the research community has witnessed an emergent interest in developing Large Multimodal Models (LMMs)~\cite{achiam2023gpt4,liu2024llava,chen2024internvl} that have exhibited impressive abilities in a variety of benchmarks~\cite{yue2024mmmu,liu2023mmbench}. The central purpose behind these explorations is to build human-centric AI models that can serve as helpful assistants for everyday life. Given that over 57\% of the global population resides in urban areas~\cite{worldbank2024urbanpopulation}, it is crucial that these AI models should be capable of performing a variety of urban tasks, such as assisting government officials to manage urban development and facilitating citizens to make decisions in daily life~\cite{zhou2024llm4urbanplanning,feng2024citygpt}. On the other hand, urban areas are often captured from various perspectives, including the vertical view from satellite or aerial imagery and the horizontal view from street-view imagery. 
To be truly effective in assisting the large urban population, AI models should also be capable of comprehensively understanding these environments from multiple perspectives.
 
To better evaluate and develop human-centric AI models, several works have explored the prospect of LMMs for urban environments. For example, various works evaluate the capabilities of LMMs on region-level visual recognition~\cite{hao2024urbanvlp,yan2024urbanclip} and urban planning~\cite{zhou2024llm4urbanplanning}. Besides, researchers also examine the performance of LMMs with remote sensing images~\cite{li2024vrsbench,kuckreja2024geochat}. However, as shown in Fig.\ref{fig:question-comparison}, while these studies primarily focus on urban understanding at a region level, they neglect human-centric tasks within the urban scenarios. A more comprehensive approach should address urban tasks across multiple levels, from region-level recognition tasks to role-level tasks such as geo-localization and scene understanding.

Another important aspect of the urban environments is that they are usually captured by multiple different perspectives. As each perspective offers unique information, it is vital for LMMs to comprehend different perspectives to complete certain tasks. For instance, geo-localization tasks require satellite-view imagery for spatial orientation and street-view imagery for detailed contexts. LMMs must utilize both perspectives to successfully perform geo-localization. Given the importance of understanding urban environments from multiple perspectives, current benchmarks that only evaluate LMMs on single-view data~\cite{wang2024earthvqa,feng2024citybench}, as shown in Fig.\ref{fig:question-comparison}, are incomprehensive. Therefore, it is crucial to develop a multi-view benchmark to accurately evaluate LMMs under complex urban settings. However, one of the key challenges of curating such datasets is create annotations for cross-view scenarios~\cite{zhu2021vigor,ye2024sg}. While paired street and satellite images are easy to acquire, creating questions about cross-view relations remains challenging due to lack of annotations~\cite{li2023omnicity}.

To address these challenges, we propose UrBench, a multi-task, multi-view benchmark designed for comprehensively evaluating LMMs in urban environments. UrBench comprises over 11.6K questions across 14 tasks spanning four dimensions: Geo-Localization, Scene Understanding, Scene Reasoning, and Object Understanding. UrBench includes both region-level tasks from previous benchmarks as well as role-level tasks aimed at assisting humans in daily life. Additionally, considering the multi-view characteristics of urban environments, UrBench incorporates multiple urban perspectives to evaluate LMMs' capabilities in understanding complex multi-view relations. 
In constructing UrBench, we introduce a novel cross-view detection and matching method to create multi-view annotations. We then utilized these images and annotations to construct our high-quality and diverse set of questions with various methods. 
As shown in Fig.\ref{fig:top4_model}, our evaluation results indicate that LMMs lag behind human experts in most tasks, highlighting their limitations towards human-centric assistants in urban environments. Our contributions are summarized as follows:

\begin{figure}[t]
    \centering
    \includegraphics[width=1\linewidth]{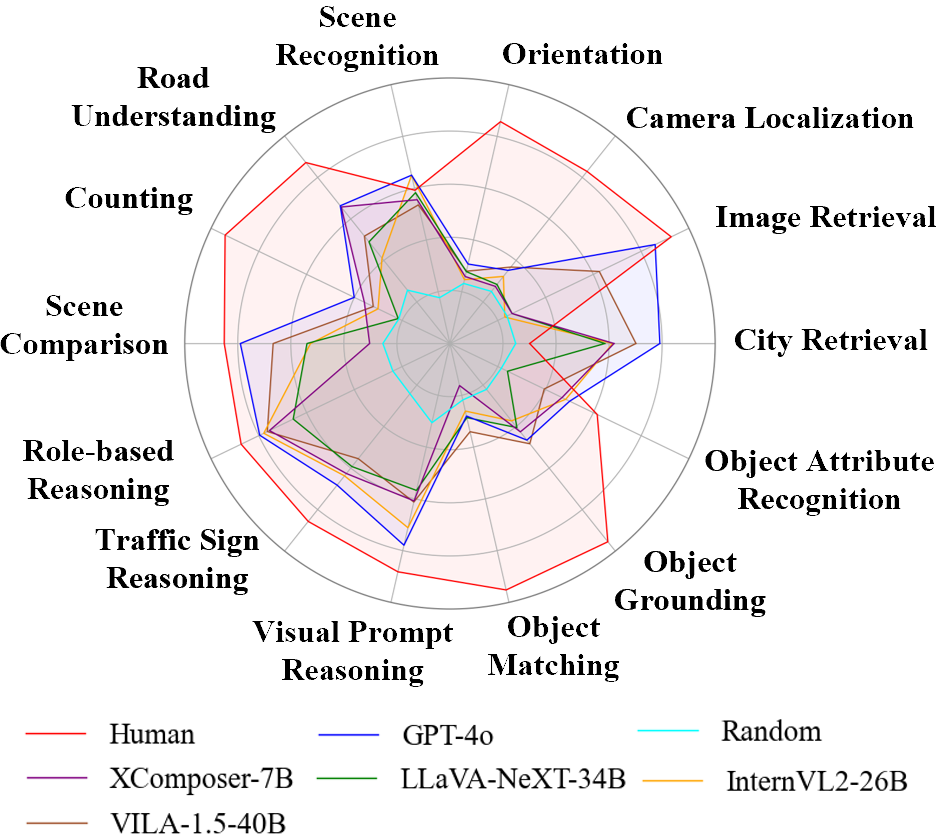}
    \caption{The performances of the 5 leading LMMs, as well as that of the human and random guess, on UrBench.}
    \label{fig:top4_model}
\end{figure}

\begin{itemize}
    \item We propose UrBench, a multi-view benchmark designed to evaluate LMMs' performances in urban environments. Our benchmark includes 14 urban tasks that we categorize into various dimensions. These tasks encompass both region-level evaluations that assess LMMs' capabilities in urban planning, as well as role-level evaluations that examine LMMs' responses to daily issues.

    \item We introduce a novel benchmark curation pipeline that involves a cross-view detection-matching algorithm for object-level annotation generation and a question generation approach that integrates LMM-based, rule-based, and human-based methods. This pipeline ensures the creation of a large-scale and high-quality corpus of questions, significantly enhancing the diversity and depth of evaluation across multiple urban tasks.

    \item We evaluate 21 popular LMMs on UrBench. Our evaluation results show that current models lag behind human experts in most tasks and reveal LMMs' inconsistent behaviors with different urban views, which demonstrates the limitations of current LMMs in urban environments.
    
\end{itemize}

\section{Related Work}\label{sec:relatedwork}

\begin{figure*}[htbp]
    \centering
    \hspace{0.1cm}
    \begin{subfigure}{0.35\textwidth}
        \centering
        \includegraphics{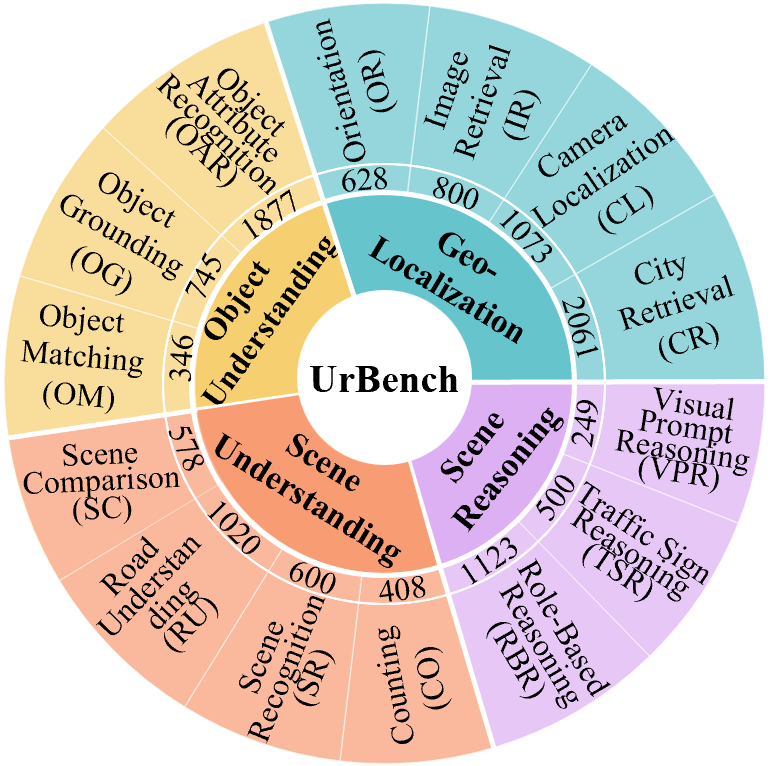}
        \caption{}
        \label{fig:sub1}
    \end{subfigure}
    \hspace{-0.2cm}
    \begin{subfigure}{0.35\textwidth}
        \centering
        \includegraphics{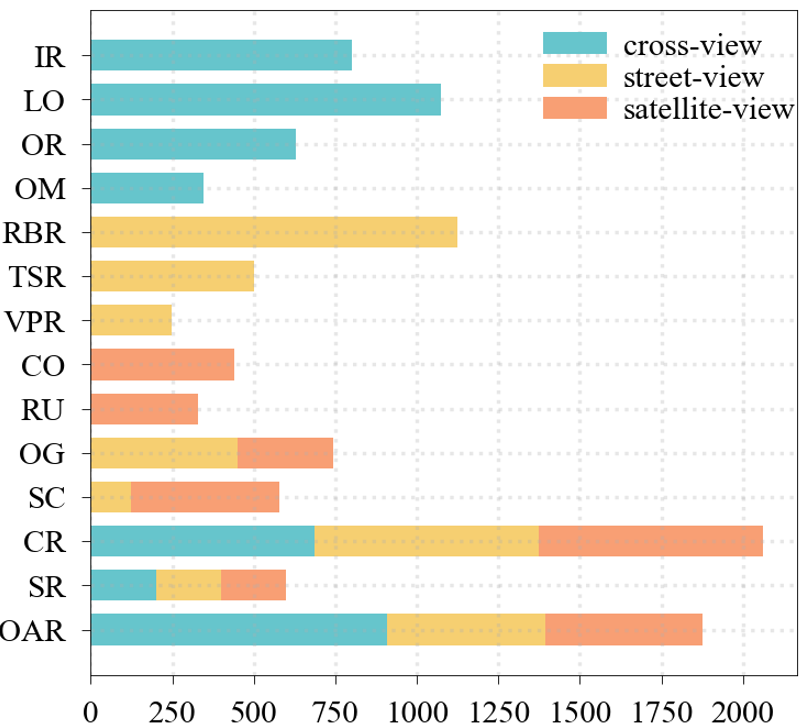}
        \caption{}
        \label{fig:sub2}
    \end{subfigure}
    \hspace{-0.1cm}
    \begin{subfigure}{0.28\textwidth}
        \centering
        \includegraphics[width=\textwidth]{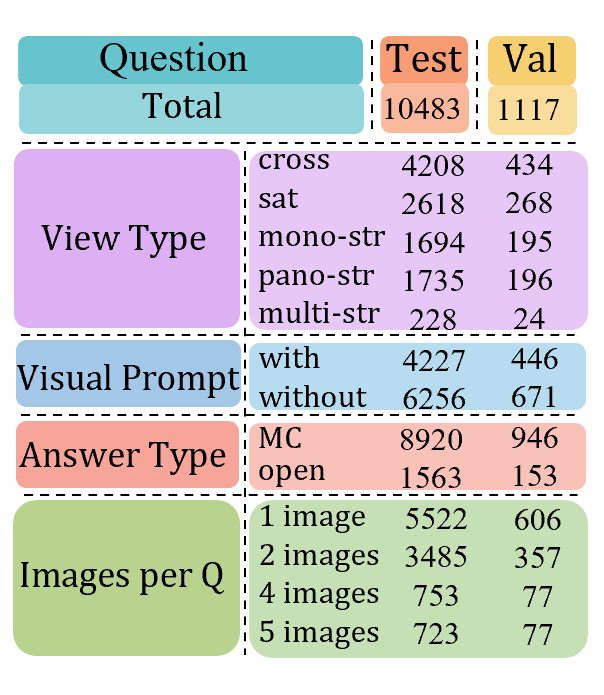}
        \caption{}
        \label{fig:sub3}
    \end{subfigure}
    \caption{(a) The 14 types of tasks under 4 evaluation dimensions. (b) The view types of each task. (c) The statistics of UrBench. cross, sat, and str are the abbreviations for cross-view, satellite-view, and street-view. mono, pano, and multi are the abbreviations of monocular, panoramic, and multiple. MC and open means multiple-choice and open-ended, respectively.}
    \label{fig:statistics}
\end{figure*}

\subsection{Large Multimodal Models}

Building on the strengths of Large Language Models (LLMs)~\cite{brown2020gpt3, touvron2023llama} in complex language reasoning and understanding, Large Multimodal Models (LMMs) can process inputs from multiple modalities and accomplish sophisticated visual reasoning and understanding tasks~\cite{yin2023survey,bai2024survey}. The rapid growth of LMMs has given rise to both closed-source models like GPT-4o~\cite{achiam2023gpt4} and Gemini~\cite{reid2024gemini}, as well as open-source models such as LLaVA~\cite{liu2024llava}, and VILA~\cite{lin2024vila}, all of which have demonstrated significant potential for various application tasks~\cite{cui2024survey,xiao2024comprehensive}.

For urban-related tasks, recent developments show growing interest in utilizing LLMs and CLIP~\cite{radford2021clip}, covering aspects such as urban planning and vision-language navigation~\cite{zhou2024llm4urbanplanning,schumann2024velma}.
For instance, UrbanCLIP~\cite{yan2024urbanclip} leverages LLMs~\cite{touvron2023llama} and CLIP~\cite{radford2021clip} for urban region profiling using remote sensing images, while Velma~\cite{schumann2024velma} combines LLMs with CLIP~\cite{radford2021clip} for street view navigation.  Scene-LLM leverages LLMs for multi-view 3D reasoning in in-door setups, while CityGPT~\cite{feng2024citygpt} studies LLMs' performances in urban spatial understanding tasks.


\subsection{Multimodal Benchmarks}
With the rapid advancement of LMMs, traditional multimodal question answering benchmarks like VQA~\cite{goyal2017vqa} and GQA~\cite{hudson2019gqa} have become inadequate for fully assessing LMM capabilities. Recently, more comprehensive benchmarks are introduced to better evaluate LMMs. For example, MME~\cite{fu2023mme} is one of the first to thoroughly assess LMMs across 14 perception and reasoning tasks. MMMU~\cite{yue2024mmmu} benchmarks expert-level knowledge using college-level questions, showing that current models still lag behind human experts. Newer benchmarks, such as those in~\cite{jiang2024mantis, wang2024muirbench}, focus on multi-image reasoning, and MUIRBench~\cite{wang2024muirbench} includes tasks with unanswerable questions. While these works extensively evaluate LMMs in visual reasoning and multi-image understanding, few analyze performance in urban environments. Our work fills this gap by constructing a comprehensive benchmark for evaluating LMMs' potential in urban planning, reasoning, and understanding from multi-views.

There is also a growing body of work focused on benchmarking LMMs in the remote sensing domain. Early efforts like RSVQA~\cite{lobry2020rsvqa} comprises visual recognition tasks such as classification and detection for image sensing images. EarthVQA~\cite{wang2024earthvqa} feature remote sensing image-question pairs centered on attributes of ground objects in urban areas. RSIEval~\cite{hu2023rsgpt} and LHRS-Bench~\cite{muhtar2024lhrs} adapt existing remote sensing datasets to create visual reasoning benchmarks for LMMs, while Geochat~\cite{kuckreja2024geochat} primarily assesses regional perception capabilities. VRSBench~\cite{li2024vrsbench} utilizes GPT-4~\cite{achiam2023gpt4} to generate visual question answering data focused on object relations. More recently, CityBench~\cite{feng2024citybench} evaluates LMMs in urban environments but with a limited range of tasks.
Overall, these existing benchmarks are limited in task variety and lack multi-view samples. Moreover, because the image data in these works are primarily repurposed from existing datasets~\cite{xia2018dota,sun2022fair1m}, their geographical diversity is constrained. In contrast, our benchmark introduces a new pipeline for collecting multi-view and multi-image data, expanding the range of tasks and incorporating images from diverse geolocations with multiple perspectives.

\section{UrBench}\label{sec:construct}

\subsection{Benchmark Analysis}

\begin{figure*}
    \centering                         
    \includegraphics{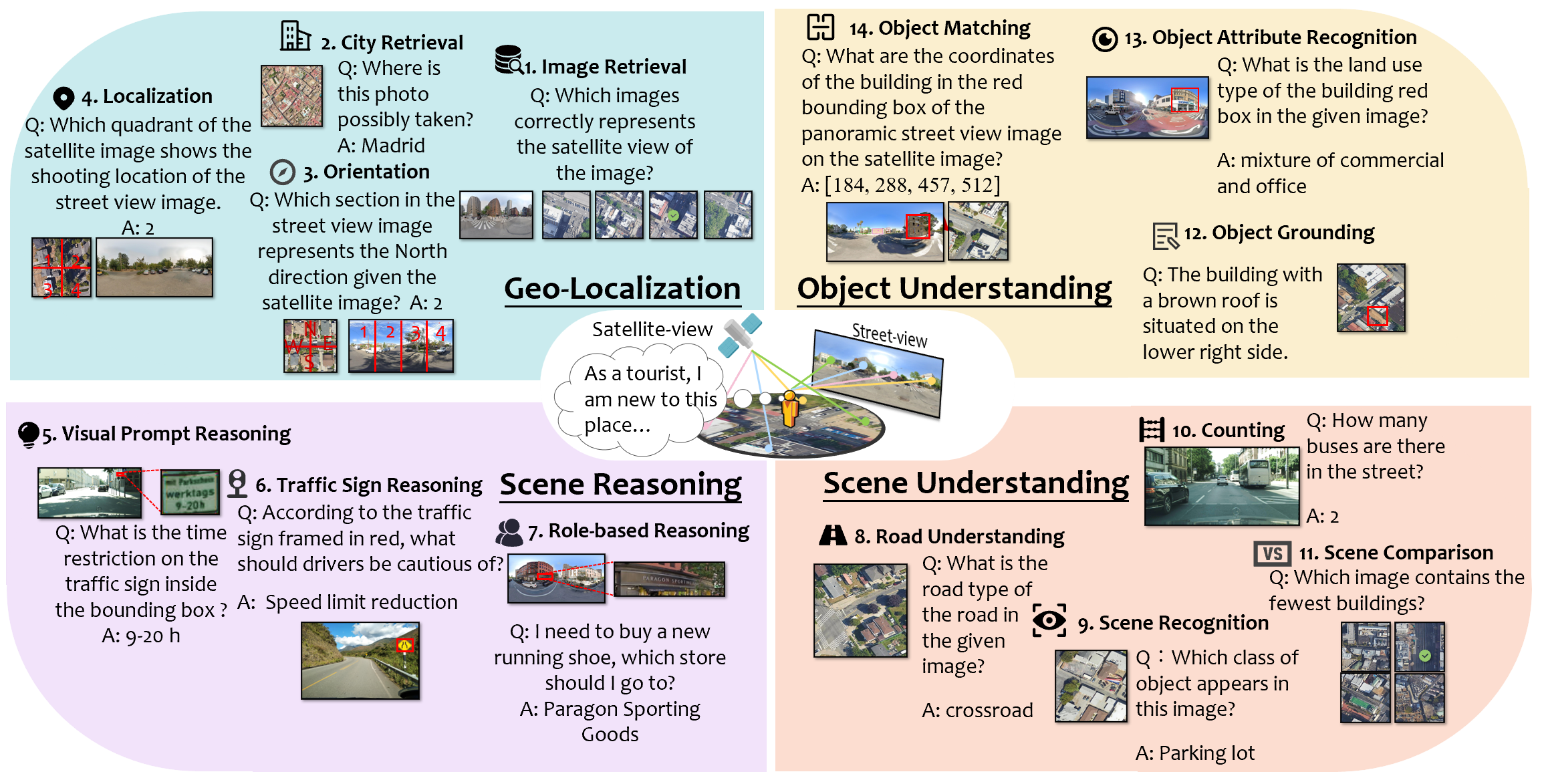}
    \caption{UrBench consists of 14 different task types, categorized into four evaluation dimensions based on the capacities and the granularity of the objects of interest assessed by the questions.}
    \label{fig:tasks}
\end{figure*}

\textbf{Overview.} We introduce UrBench, a novel benchmark designed for evaluating LMMs in urban scenarios. 
As detailed in Fig.\ref{fig:statistics}(c), UrBench comprises 11.6K questions, which are divided into a validation set for hyperparameter selection and a test set for evaluation. The validation set and the test set contain approximately 1.1K and 10.5K questions, respectively. Please refer to the appendix for further statistical details.
The UrBench is characterized by the following features:
(1) UrBench integrates street-view, satellite-view, and street-satellite cross-view images, offering a more comprehensive understanding of urban scenarios (Fig.\ref{fig:statistics}(b)).
(2) UrBench evaluates the capability of LMMs focusing on urban scenarios from comprehensive dimensions, including Geo-Localization, Scene Reasoning, Scene Understanding, and Object Understanding, with a total of 14 task types (Fig.\ref{fig:statistics}(a)). 
(3) The questions of UrBench are generated by an integrated approach, encompassing model-based, rule-based, and human-based methods, which ensures the generation of a substantial and high-quality corpus of questions.

\noindent\textbf{Comparison with existing benchmarks.} 
While general benchmarks such as MUIRBench~\cite{wang2024muirbench} and MMMU~\cite{yue2024mmmu} evaluate the capacities of LMMs in general scenarios, typically from a single view, our benchmark is focused on urban scenarios from more different perspectives.
On the other hand, unlike existing benchmarks in urban scenarios such as CityBench~\cite{feng2024citybench} and EarthVQA~\cite{wang2024earthvqa} that place significant emphasis on single-view images and a limited range of tasks, our benchmark incorporates questions that utilize multi-view images and cover more diverse task types.

\begin{figure*}[t]
    \centering
    \includegraphics{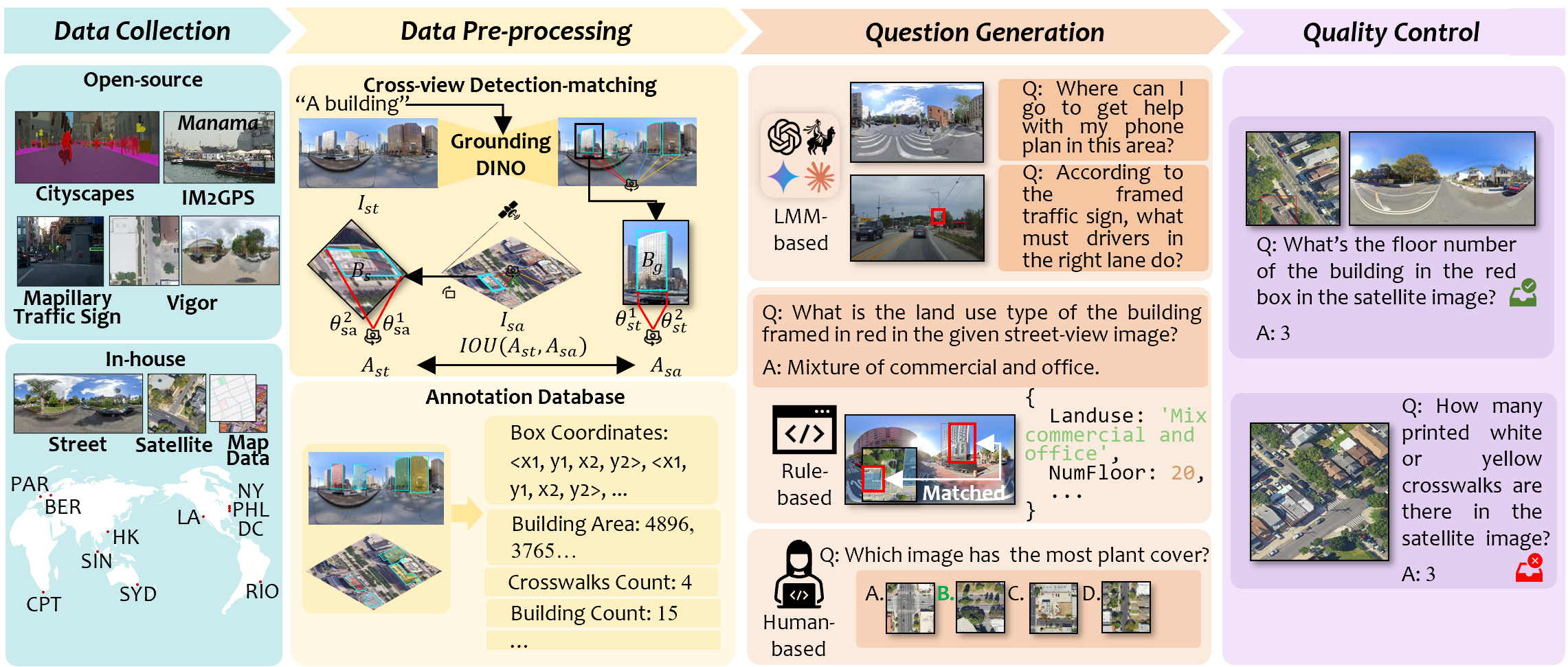}
    \caption{UrBench curation pipeline includes data collection, data pre-processing, question generation and quality control. 
    }
    \label{fig:data_pipeline}
\end{figure*}

\subsection{Benchmark Tasks}
UrBench comprehensively evaluates LMMs in urban scenarios from four evaluation dimensions. Fig.\ref{fig:tasks} illustrates the specific task types under each evaluation dimension. Please refer to the appendix for additional task questions.


\noindent\textbf{Geo-Localization.} This dimension contains role-level tasks widely used in remote-sensing~\cite{zhu2021vigor}, which requires LMMs to predict geographical coordinates and directions given images (Task 1-4 in Fig.\ref{fig:tasks}). For example, in Image Retrieval (IR), we query LMMs with satellite or street view images to retrieve their corresponding counterpart, while City Retrieval (CR) tasks LMMs to predict the name of a city given its satellite or street view images. Orientation (OR) and Camera Localization (CR) require LMMs to pinpoint directions given cross-view information. 


\noindent\textbf{Scene Reasoning.} Reasoning about urban scenes is crucial for assisting humans in urban environments. In this dimension, we design three role-level tasks that aim to assess the reasoning capabilities of LMMs under urban multi-view scenarios (Task 5-7 in Fig.\ref{fig:tasks}). In Visual Prompt Reasoning (VPR), LMMs must reason about objects framed in visual prompts, while in Traffic Sign Reasoning (TSR), LMMs predict the usage of traffic signs. To better simulate real-world human usage, we design Role-based Reasoning to incorporate questions that are posed from the perspective of urban residents, such as shoppers, visitors, and city managers.

\noindent\textbf{Scene Understanding.} To better evaluate LMMs' capabilities in region-level scene understanding, we design new tasks as well as adapting existing tasks to urban environments in this dimension, as shown in (Task 8-11 in Fig.\ref{fig:tasks}). For example, to examine whether LMMs can detect common urban regions like crosswalks, we convert the classic Counting (CO)~\cite{ranjan2021count} task to our specific settings. We design Scene Recognition (SR) and Scene Comparison (SC) to assess LMMs' region-level understanding of urban scenes, such as recognizing building types and plant cover. Additionally, we present Road Understanding (RU), which is aimed at examining LMMs' ability in classifying road types and understanding traffic roads.


\noindent\textbf{Object Understanding.} In Object Understanding, we assess the fine-grained region-level capabilities of LMMs within urban environments (Task 12-14 in Fig.\ref{fig:tasks}). Inspired by previous grounding tasks~\cite{yu2016modeling}, we include Object Grounding (OG) to evaluate LMMs' abilities to ground text phrases of  objects in images across different views. To further challenge the spatial understanding of LMMs, we propose Object Matching (OM), a cross-view task where LMMs predict the locations of objects in satellite and street views based on their cross-view correspondences. Object Attribute Recognition (OAR) prompts LMMs to predict ground object attributes such as building floors and land use.

\begin{table*}[t]
  \centering
  \fontsize{7pt}{4pt}\selectfont
  \setlength{\tabcolsep}{1.5mm}{
    \begin{tabular}{c|cccc|cccc|ccc|ccc|c}
    \toprule
    \multirow{1}[4]{*}{\textbf{Model}} 
    & \multicolumn{4}{c|}{\shortstack{\textbf{Geo-} \\ \textbf{Localization}}}
    & \multicolumn{4}{c|}{\shortstack{\textbf{Scene} \\ \textbf{Understanding}}}
    & \multicolumn{3}{c|}{\shortstack{\textbf{Scene} \\ \textbf{Reasoning}}}
    & \multicolumn{3}{c|}{\shortstack{\textbf{Object} \\ \textbf{Understanding}}} 
    & \multirow{1}[4]{*}{\textbf{Overall}} \\
\cmidrule{2-15}          & \textbf{CR} & \textbf{IR} & \textbf{CL} & \textbf{OR} & \textbf{SR} & \textbf{RU} & \textbf{CO} & \textbf{SC} & \textbf{RBR} & \textbf{TSR} & \textbf{VPR} & \textbf{OM} & \textbf{OG} & \textbf{OAR} &  \\
    \midrule
    Human & 30.0  & 92.6  & 82.9  & 85.7  & 59.2  & 87.2  & 94.1  & 85.1  & 87.4  & 85.7  & 88.2  & 95.2  & 95.5  & 61.6  & 69.9  \\
    Random & 24.8  & 23.9  & 25.1  & 23.2  & 17.7  & 25.7  & 21.4  & 25.3  & 23.9  & 24.2  & 30.6  & 21.8  & 22.1  & 21.5  & 23.5  \\
    \midrule
    GPT-4o & \textbf{79.2 } & \textbf{85.9 } & \underline{35.3}  & \underline{30.7}  & \textbf{65.0 } & \underline{66.3}  & 40.1  & \underline{79.0}  & \textbf{79.6 } & 68.2  & 77.9 & 28.0  & 46.5  & \underline{50.1} & \textbf{61.2} \\
    Gemini-1.5-Flash & 69.7  & 25.9  & 25.9  & 24.0  & 57.9  & \textbf{71.0}  & 29.1  & 67.7  & 77.8  & 75.8 & 69.8  & 22.0  & 39.1  & 40.9  & 50.9  \\
    Claude-3.5-Sonnet & \underline{72.3}  & 55.8  & 30.8  & \textbf{33.3}  & 52.4  & 59.0 & \textbf{48.0 } & \textbf{81.0 } & 73.7  & 37.7  & 66.7  & 22.0  & \textbf{61.5 } & 45.4  & \underline{55.0}  \\
    \midrule
    TinyLLaVA & 51.9  & 23.2  & 24.7  & 27.9  & 8.6   & 9.3  & 9.5  & 27.6  & 40.3  & 32.7  & 48.6  & 22.9  & 41.1  & 18.8  & 29.9  \\
    InternVL2-2B & 50.3  & 23.8  & 31.9  & 29.0  & 47.9  & 47.9  & 28.6  & 30.1  & 64.8  & 45.9  & 54.5  & 25.5  & 30.3  & 40.5  & 41.2  \\
    InternVL2-4B & 55.0  & 24.2  & 27.4  & 23.1  & 52.3  & 53.1  & 22.1  & 39.4  & 73.0  & 56.6  & 62.6  & 30.6  & 30.2  & 40.2  & 43.5  \\
    XComposer2-4KHD & 61.9  & 26.0  & 27.5  & 25.9  & 55.6  & 65.8  & 35.8  & 30.3  & 75.5  & 62.6  & 60.8  & 16.2  & 42.6  & 46.9  & 47.8  \\
    LLaVA-NeXT-7B-Mistral & 51.6  & 25.9  & 24.2  & 24.0  & 55.6  & 47.3  & 34.1  & 28.4  & 59.2  & 42.7  & 45.5  & \textbf{43.9} & 33.1  & 30.0  & 39.2  \\
    LLaVA-NeXT-7B-Vicuna & 51.2  & 24.6  & 27.2  & 23.3  & 56.1  & 20.2  & 34.3  & 25.9  & 49.6  & 51.5  & 54.5  & 31.8  & 27.2  & 31.9  & 37.1  \\
    InstructBLIP-Vicuna-7B & 40.4  & 25.7  & 25.4  & 27.1  & 33.0  & 14.8  & 20.4  & 28.0  & 30.7  & 25.7  & 29.3  & 22.9  & 22.7  & 17.3  & 27.6  \\
    LLaVA-NeXT-Interleave-7B & 57.9  & 41.6  & 27.6  & 25.5  & 52.6  & 50.7  & 37.3  & 48.4  & 65.8  & 55.9  & 63.1  & 37.3  & 37.6  & 41.5  & 40.4  \\
    Mantis-LLaMA3-SigLIP & 67.0  & 32.4  & 27.0  & 27.2  & \underline{59.2}  & 44.5  & 27.4  & 52.4  & 67.6  & 41.6  & 57.7  & 25.2  & 34.2  & 38.6  & 45.3  \\
    Mantis-Idefics2 & 69.0  & 29.9  & 27.0  & 25.7  & 50.3  & 49.3  & 22.9  & 56.0  & 68.9  & 50.6  & 56.3  & 29.6  & 37.8  & 35.6  & 40.7  \\
    LLaVA-NeXT-8B & 54.4  & 27.0  & 27.8  & 26.0  & 55.7  & 44.5  & 34.1  & 24.0  & 55.2  & 52.8  & 58.6  & \underline{39.8}  & 41.7  & 28.4  & 38.8  \\
    InternVL2-8B & 50.8  & 26.6  & 31.8  & 25.2  & 53.0  & 52.6  & \underline{43.0}  & 51.4  & 74.9  & 54.8  & 62.6  & 30.3  & 32.0  & 41.3  & 48.8  \\
    Idefics-2-8B & 65.5  & 23.8  & 26.0  & 24.1  & 52.3  & 47.9  & 25.9  & 27.8  & 64.7  & 60.4  & 42.8  & 24.8  & 21.1  & 27.4  & 42.7  \\
    LLaVA-NeXT-13B & 52.0  & 24.5  & 27.7  & 26.7  & 53.9  & 50.7  & 33.8  & 25.1  & 54.0  & 52.1  & 52.3  & 31.8  & 34.8  & 26.3  & 46.5  \\
    VILA-1.5-13B & 62.7  & 33.7  & 28.6  & 24.1  & 47.7  & 43.9  & 23.9  & 48.6  & 66.3  & 43.8  & 46.4  & 25.8  & 32.3  & 38.2  & 45.8  \\
    InternVL2-26b & 61.3  & 23.0  & 32.3  & 24.7  & \textbf{65.0 } & 41.1  & 30.1  & 52.6  & 77.9  & 63.8  & 71.2  & 26.1  & 37.3  & 48.4  & 46.0 \\
    LLaVA-NeXT-34B & 58.4  & 26.0  & 28.5  & 27.8  & 58.3  & 49.0  & 21.6  & 53.9  & 65.6  & 59.3  & 56.8  & 28.7  & 40.5  & 24.1  & 43.7  \\
    VILA-1.5-40B & 70.1  & \underline{62.5}  & \textbf{36.8 } & 27.9  & 53.6  & 51.7  & 32.1  & 66.7  & 76.4  & 55.5  & 61.3  & 34.1  & \underline{48.3}  & 39.5  & 53.1  \\
    \bottomrule
    \end{tabular}%
    }
    \caption{The quantitative results for 3 closed-source and 18 open-source LMMs, as well as those for human and random guess across 14 tasks. The overall score is computed across all tasks. The maximum value and the next largest value of each task are indicated by the \textbf{bold} and \underline{underlined} text, respectively. Task names are abbreviated for brevity.}
  \label{tab:main-table}%
\end{table*}%

\subsection{Benchmark Curation}

\noindent\textbf{Data Collection.} As outlined in the data collection stage in Fig.\ref{fig:data_pipeline}, there are two data sources of UrBench, the in-house data collected by ourselves and data from open datasets. 
The in-house data contains 2,604 street-view images from Google Street View and 4,239 satellite-view images from Google Earth (Level 19). Among these images, 1,965 street-satellite image pairs are fit together according to their geological coordinates. In addition, each satellite-view image is equipped with some ground object annotation from OpenStreetMap\footnote{https://www.openstreetmap.org}. 
We follow previous works~\cite{zhu2021vigor} to ensure no significant temporal differences between satellite and street-view images, which were all collected in 2022-2023.
To support more urban tasks, we also collect images from existing open source datasets, including Cityscapes~\cite{Cordts2016Cityscapes}, Mapillary Traffic Sign Dataset~\cite{ertler2020mapillary}, VIGOR dataset~\cite{zhu2021vigor}, and IM2GPS~\cite{hays2008im2gps}.


\noindent\textbf{Data Pre-processing.} In this stage, we process our collected raw image data to produce annotations for later stages. For cross-view tasks that require object matching, we develop a cross-view detection-matching method to obtain paired-up instance level annotations. 
Specifically, We first use a pretrained Grounding DINO~\cite{liu2023grounding} to obtain bounding box annotations for street-view images. Since bounding box annotations for satellite-view images are already obtained through OSM, we apply ray tracing to map the street-view boxes to the satellite view. Next, we calculate the IoUs between the mapped and original satellite-view boxes and select pairs with IoUs over 0.5 as cross-view matches. We then employ human checking to further ensure the quality of our matching algorithm. This way, we effectively align bounding boxes from different views at an instance level.
Additionally, we construct a comprehensive annotation database by unifying annotations from different datasets, facilitating data generation and quality control.


\noindent\textbf{Question Generation.} Given the nature and requirements of different tasks, we design three methods to generate question samples for UrBench.
(1) LMM-based. For Scene Reasoning tasks where questions cannot be derived from our annotations, we prompt LMMs to generate Q\&A pairs based on specific task settings. To mitigate bias from LMMs generating and testing their own questions, we diversify our samples by using four different LMMs: GPT-4~\cite{achiam2023gpt4}, Gemini-1.5-Flash~\cite{reid2024gemini}, Claude-3.5-Sonnet~\cite{anthropic2023claude}, and InternVL2-26B~\cite{chen2024internvl}. These samples are then reviewed by humans to eliminate potential hallucinations and ensure quality.
(2) Rule-based. For tasks with fixed settings, such as IR and CL, we generate questions using rule-based templates given the image annotations, which allows us to automatically convert the annotations into corresponding Q\&A pairs, ensuring consistency and efficiency.
(3) Human-based. For tasks like SC where answers cannot be derived from annotations, we have human annotators determine the ground truths and generate the Q\&A pairs.


\noindent\textbf{Quality Control.} At this stage, we employ human checking to alleviate potential bias during construction of UrBench. 
Specifically, for cross-view images, to minimize temporal inconsistencies, annotators are asked to remove images with significant temporal changes. 
During the preprocessing stage, to reduce potential bias introduced by our cross-view detection-matching method, human verification ensures the correctness of paired bounding boxes and filters out mismatched samples.
Furthermore, for LMM-generated data, multiple annotators are engaged to eliminate hallucinations and maintain data quality.
These comprehensive quality control steps ensure the robustness and accuracy of UrBench.

\section{Experiments}\label{sec:experiments}

In this section, we evaluate various LMMs on our proposed UrBench. We consider closed-source models, open-source single-image models and open-source multi-image models, and perform evaluations under zero-shot settings. 
In the following sections, we first introduce our evaluated models evaluation protocols. Then we summarize our findings of model performance with respect to different model types, view settings and tasks. Finally, we provide a detailed analysis in terms of different tasks and views.

\subsection{Evaluation Setups}\label{sec:eval-setups}

\textbf{Evaluated Models.} We evaluate 3 closed-source and 18 open-source LMMs across different model types and sizes. For closed-source models, we consider GPT-4o~\cite{achiam2023gpt4}, Gemini-1.5-Flash~\cite{reid2024gemini}, Claude-3.5-Sonnet~\cite{anthropic2023claude}. For open-sourced models, we categorize them into single-image type and multi-image type according to their training data and strategies, including LLaVA series~\cite{liu2024llavanext}~\cite{zhou2024tinyllava}, XComposer~\cite{zhang2023internlm-xcomposer}, InstructBLIP~\cite{li2023blip2} and idefics~\cite{laurenccon2024idefics} for single-image type, and Mantis series~\cite{jiang2024mantis}, VILA series~\cite{lin2024vila}, InternVL series~\cite{chen2024internvl} and LLaVA-NeXT-Interleave\cite{li2024llava-next-interleave} for multi-image type.

\noindent\textbf{Evaluation Protocols.} In UrBench, our questions have two response formats: multiple-choice and open-ended. We follow standard setups in MMMU~\cite{yue2024mmmu} and MuirBench~\cite{wang2024muirbench} to process LMMs' responses. To ensure reproducibility, we set the temperature to 0 and perform greedy decoding. Additionally, for models that do not support multi-image inputs, we concatenate the images as one input. More details on setups and the human evaluation protocols are provided in the Appendix.

\subsection{Main Results}\label{sec:main-results}


\noindent{\textbf{Overall Challenge Presented in UrBench. }}As indicated by Table \ref{tab:main-table}, UrBench poses significant challenges to current SoTA LMMs. We find that the best performing closed-source model GPT-4o and open-source model VILA-1.5-40B only achieve a 61.2\% and a 53.1\% accuracy, respectively. Interestingly, our findings indicate that the primary limitation of these models lies in their ability to comprehend UrBench questions, not in their capacity to process multiple images, as the performance between multi-image and their single-image counterparts shows little difference, such as LLaVA-NeXT-8B and LLaVA-NeXT-Interleave in Table \ref{tab:main-table}. Overall, the challenging nature of our benchmark indicates that current LMMs' strong performance on the general benchmarks~\cite{fu2023mme,liu2023mmbench} are not generalized to the multi-view urban scenarios.

\noindent{\textbf{LMMs' performances across task dimensions. }} In Table \ref{tab:main-table}, we show LMMs performances across different dimensions. We observe that most LMMs exhibit impressive capabilities in Scene Reasoning tasks such as Visual Prompt Reasoning (VPR) and Role-based Reasoning (RBR), which are greatly aligned with their SFT objectives~\cite{liu2024llava}. 
In Scene Understanding, models perform relatively well in region-level tasks such as Scene Recognition (SR) and Scene Comparison (SC), but are bad at Counting (CO). 
While LMMs achieve impressive results in City Retrieval (CR), however, in other Geo-localization tasks, e.g., Camera Localization (CL) and Orientation (OR), most LMMs only perform slightly better or worse than random guessing, yielding a 28.6\% and a 26.3\% average accuracy, respectively. 
Overall, LMMs' exception capbilities in reasoning tasks and world knowledge are well-examplified in our benchmark. However, our benchmark also demonstrates current models' limited abilities in handling other important urban tasks such as geo-localization.

\begin{figure}[t]
    \centering
    \includegraphics{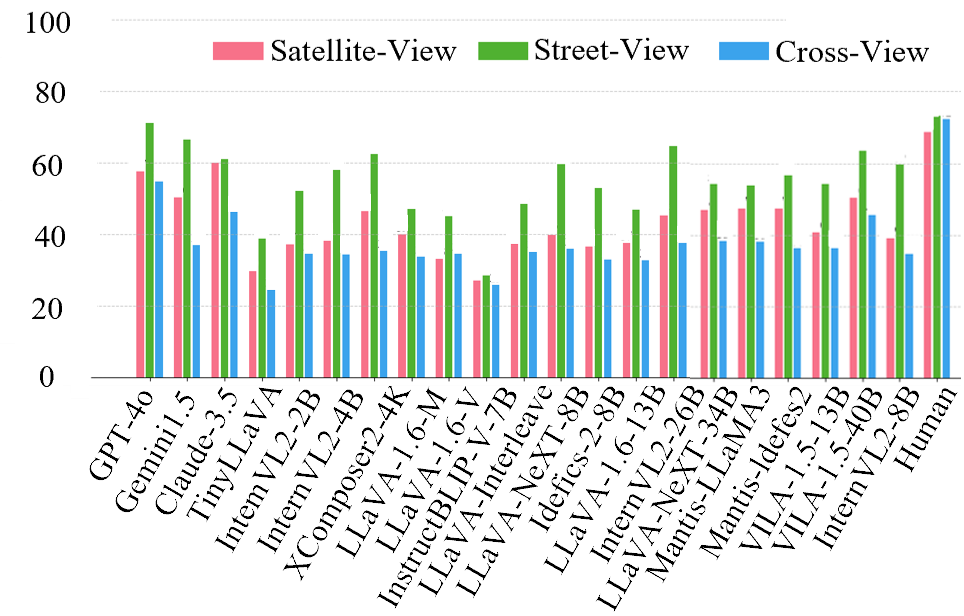}
    \caption{Quantitative result comparison across 3 views. }
    \label{fig:model_performance_bar}
\end{figure}

\noindent{\textbf{Are LMMs consistent with Different Views?}} Out of the three views in our proposed benchmark, we find that models struggle the most with cross-view tasks, averaging only 36.2\% in all models, while averaging 54.6\% and 42.3
\% in street-view and satellite-view, respectively. Fig.\ref{fig:model_performance_bar} presents the overall model performance across different views. Claude-3.5-Sonnet~\cite{anthropic2023claude} obtains the highest 60.1\% score in satellite-view. GPT-4o~\cite{achiam2023gpt4} achieves the highest average score in street-view tasks, with Gemini-1.5-Flash~\cite{reid2024gemini}, InternVL2-26B~\cite{chen2024internvl} and VILA-1.5-40B~\cite{lin2024vila} following close. Most models perform no better than 40\% in cross-view tasks, except for GPT-4o~\cite{achiam2023gpt4}, Claude-3.5-Sonnet~\cite{anthropic2023claude} and VILA-1.5-40B~\cite{lin2024vila}. Our results show that current LMMs are best at street-view tasks, while handling satellite-view and cross-view tasks insufficiently. Future work could incorporate more diverse and relevant training data with multiple views to help LMMs understand urban scenes holistically.

\noindent{\textbf{Does GPT-4o Surpass Human Experts? }} Human experts outperform GPT-4o by an average of 17.4\% and achieve better performance in 12 out of the 14 UrBench tasks. Humans only fall significantly behind GPT-4o in the City Retrieval task, which we attribute to the rich geographic world knowledge in the LMMs. For humans, identifying the correct geographic location from a photo is challenging. GPT-4o falls behind human experts by 54.1\% , on simple tasks such as counting, and 67.8\% on Object Grounding (OG). We note these gaps highlight the significant room for improvement in LMMs' urban understanding capabilities.


\noindent{\textbf{Disparity between Closed-source and Open-source Models.}} Under our urban settings, we observe that the gap between closed-source and open-source models are closing in. While leading open-source models like VILA-1.5-40B~\cite{lin2024vila} is still behind the leading closed-source model GPT-4o, VILA-1.5 demonstrates superior performance over Gemini-1.5-Flash and is close with Claude-3.5. Our results show the potential of open LMMs as urban assistants.


\subsection{Detailed Analysis}\label{sec:detailed analysis}

\noindent{\textbf{LMMs struggle to understand cross-view relations.}} Several UrBench tasks involve understanding the internal relations between satellite and street view images. While prior works~\cite{shi2022cross-view-retrieval,ye2024cross} have demonstrated that specialist models can achieve impressive results in cross-view tasks such as Camera Localization (CL) and Orientation (OR), our results indicate that general LMMs only possess very limited cross-view understanding capability, where their performances in average are only 3\% higher than random. Although LMMs are capable of understanding relations across multiple images~\cite{jiang2024mantis}, our findings show that their capabilities are yet to generalize to images across views. One potential direction for future work is to explore multi-view pretraining that aligns multi-view images with text, which enables LMMs to better process and understand cross-view information~\cite{lin2024vila}.


\noindent{\textbf{LMMs are inconsistent with different views of the same geolocation. }}Even though prompted with the same question at the same geolocation, LMMs behave differently with different views. In City Retrieval (CR), we find that models perform better in street-view (62.9\%) and cross-view (63.8\%) compared to satellite-view (50.9\%) in average. We conjecture it is because most models are not well-trained on satellite samples and the parametric knowledge of geolocation is more aligned with street-view images. However, we find that the results of Scene Recognition SR in satellite-view are much higher than the other views, as recognizing ground objects from a vertically upward view is easier than from a horizontal view. The experiments exhibit the imbalance and bias between views during the training of LMMs.

\section{Conclusion}\label{sec:conclusion}
In this work, we present UrBench, a new benchmark that evaluates LMMs in the urban environments with diverse task types and view types. To create our multi-view annotations, we propose a new data collection pipeline that pairs up cross-view images at an instance level. In the end, we collect 11.6K questions that include 14 subtasks of four dimensions. We carefully evaluate 21 LMMs on our questions and show their limitations in the urban environments. We conduct extensive analysis on measuring the performance of LMMs across different view types and task types, and show that current LMMs still lag behind human experts significantly in the urban environments. We also highlight that current LMMs struggle to understand multi-view image relations and their performance under different view types are inconsistent, shedding light on the imbalance and bias between different views during LMMs training. We hope our work can provide guidance for future work in improving the capability of LMMs in urban scenarios.

\section{Acknowledgments}
This project was funded by National Natural Science Foundation of China (Grant No. 42201358) and Shanghai AI Laboratory.

\bibliography{aaai25}

\end{document}